\PassOptionsToPackage{dvipsnames,table}{xcolor}
\documentclass[sigconf=true, nonacm=true, review=false, anonymous = false]{acmart}
\usepackage{booktabs} % For formal tables
\usepackage{multirow}

\usepackage{amsmath}
\usepackage{xspace}
\usepackage{xcolor}
\usepackage{dsfont}
\usepackage{rotating}
\usepackage{makecell}

\usepackage{amsthm}
\usepackage{subcaption}
\renewcommand{\epsilon}{\varepsilon}

\usepackage{mathtools}

\begin{document}
\title{Computing Star Discrepancies with Numerical Black-Box Optimization Algorithms}

\author{Fran\c{c}ois Cl\'ement}
% \orcid{0000-0001-9206-3698}
\affiliation{
  \institution{Sorbonne Universit\'e, CNRS, LIP6}
  \city{Paris}
  \country{France}}

\author{Diederick Vermetten}
\orcid{0000-0003-3040-7162}
\affiliation{
  \institution{Leiden Institute for Advanced Computer Science}
  \city{Leiden}
  \country{The Netherlands}
}

\author{Jacob de Nobel}
% \orcid{0000-0003-1169-1962}
\affiliation{
  \institution{Leiden Institute for Advanced Computer Science}
  \city{Leiden}
  \country{The Netherlands}
}

\author{Alexandre D. Jesus}
\orcid{0000-0001-7691-0295}
\affiliation{
  \institution{University of Coimbra, CISUC, DEI}
  \city{Coimbra}
  \country{Portugal}
}

\author{Luís Paquete}
% \orcid{0000-0001-7525-8901}
\affiliation{
  \institution{University of Coimbra, CISUC, DEI}
  \city{Coimbra}
  \country{Portugal}
}

\author{Carola Doerr}
\orcid{0000-0002-4981-3227}
\affiliation{
  \institution{Sorbonne Universit\'e, CNRS, LIP6}
  \city{Paris}
  \country{France}}

% % If the default list of authors is too long for headers.
\renewcommand{\shortauthors}{F. Cl\'ement et al.}

\begin{abstract} 
The $L_{\infty}$ star discrepancy is a measure for the regularity of a finite set of points taken from $[0,1)^d$. Low discrepancy point sets are highly relevant for Quasi-Monte Carlo methods in numerical integration and several other applications. Unfortunately, computing the $L_{\infty}$ star discrepancy of a given point set is known to be a hard problem, with the best exact algorithms falling short for even moderate dimensions around 8. However, despite the difficulty of finding the global maximum that defines the $L_{\infty}$ star discrepancy of the set, local evaluations at selected points are inexpensive. This makes the problem tractable by black-box optimization approaches.

In this work we compare 8 popular numerical black-box optimization algorithms on the $L_{\infty}$ star discrepancy computation problem, using a wide set of instances in dimensions 2 to 15. We show that all used optimizers perform very badly on a large majority of the instances and that in many cases random search outperforms even the more sophisticated solvers. We suspect that state-of-the-art numerical black-box optimization techniques fail to capture the global structure of the problem, an important shortcoming that may guide their future development.

We also provide a parallel implementation of the best-known algorithm to compute the discrepancy.
\end{abstract}


\maketitle

\sloppy
\section{Introduction}
\label{sec:intro}
Discrepancy measures are designed to quantify how regularly a point set is distributed in a given space.  Among the many discrepancy measures, the most common one is the $L_{\infty}$ star discrepancy. The $L_{\infty}$ star discrepancy of a finite point set $P \subseteq [0,1)^d$ measures the worst absolute difference between the Lebesgue measure of a $d$-dimensional box anchored in $(0,\ldots,0)$ and the proportion of points that fall inside this box. 

The $L_{\infty}$ star discrepancy is especially important because of the Koksma-Hlawka inequality in numerical integration~\cite{Koksma,Hlawka}. This inequality states that the error of approximating the integral $\int_{x \in [0,1]^d} f(x)dx$ of a function $f$ by the average function value $\frac{1}{|P|}\sum_{p \in P} f(p)$ of $f$ evaluated in the points $p \in P$ is upper-bounded by a term depending only on $f$ and another term depending on the $L_{\infty}$ star discrepancy of $P$. While we cannot control the integrand $f$, the Koksma-Hlawka inequality suggests that evaluating $f$ in point sets $P$ of small discrepancy is preferable. Quasi-Monte Carlo integration is undoubtedly the main motivation to study the design of point sets with low $L_{\infty}$ star discrepancy values~\cite{DickP10}. However, such point sets are also used in numerous other applications, including the design of experiments \cite{SantnerDoE}, computer vision \cite{MatBuilder}, and financial mathematics \cite{GalFin}. In evolutionary computation, low-discrepancy point sets are considered in the context of one-shot optimization~\cite{BousquetGKTV17,BossekDKNN20}, digital art~\cite{NeumannGDN018}, and TSP solvers~\cite{NeumannG0019}.

The study of low-discrepancy sequences by mathematicians has focused on trying to obtain the optimal asymptotic order of the discrepancy of a sequence of points; see~\cite{Niederreiter92} or the more recent~\cite{DickP10}. All currently known low-discrepancy sequences have a $L_{\infty}$ star discrepancy of order $O(\log^d(n)/n)$, with $n$ being the number of points \cite{Niederreiter92}. When $d$ increases, an exponential number of points is required for this bound to be smaller than 1, giving no information on the quality of the sets if $n$ is small. Taking so many points is impossible for typical applications and thus requires the construction of low-discrepancy point \emph{sets} of a fixed size. Put differently, in contrast to the asymptotic behavior studied by the classical low-discrepancy constructions, in Computer Science and similar applications we often require sets of low $L_{\infty}$ star discrepancy value for concrete set size $n$.
Recent attempts to construct such point sets, presented for example in~\cite{Rainville13} and~\cite{ClementDP22}, require a great number of discrepancy computations, as it is often the only criterion available to differentiate candidate point sets. That is, we are not aware of (possibly low-fidelity) surrogates of the star discrepancy value. Obtaining efficient methods of computing $L_{\infty}$ star discrepancies would therefore be an invaluable asset for the construction of low-discrepancy point sets. 

Unfortunately, the problem is known to be hard to tackle exactly. More precisely, the decision version of the $L_{\infty}$ star discrepancy computation was shown to be NP-hard~\cite{NPhard} and even W[1]-hard with respect to the dimension~\cite{W1hard}. The best exact algorithm to this day runs in $O(n^{1+d/2})$~\cite{DEM}. This algorithm performs reliably well for up to a few hundred points in dimension 8, or dozens in dimension 10 (see Section~\ref{sec:DEM}). To calculate the $L_{\infty}$ star discrepancy in higher dimensions, only a heuristic is available (based on Threshold Accepting~\cite{GnewuchWW12}), reliable up to around dimension 20 (see Section~\ref{sec:TA}). One of the main issues of the heuristic is that it only provides a lower bound: throughout the heuristic, we are only evaluating local discrepancies (the maximum of which defines the star discrepancy value of the set). Therefore, there are no guarantees on the quality of the discrepancy values returned by the heuristic. A comprehensive survey of the different attempts to compute the $L_{\infty}$ star discrepancy can be found in Chapter~10 of \cite{DoerrGW14}.

 Many applications would also require discrepancy calculations in much higher dimensions, in some cases $d \ge 100$, far out of reach for the moment. Despite the complexity of finding the global maximum, local evaluations are very cheap. Black-box optimization approaches can be used to tackle this problem, even making a very large number of local evaluations is feasible as the Threshold Accepting heuristic has shown.

 \textbf{Our Contributions.} We show in this work that popular numerical black-box optimizers are unsuccessful in computing the $L_{\infty}$ star discrepancy, even for instances that can be solved within a second with naïve methods. We apply 8 optimizers on three different point set types, for varying dimensions and set sizes, and show that their performance is globally bad. The relative performance rapidly decreases with increasing dimension, whereas the size of the point sets only seems to have a minor impact on the overall (bad) quality of the solvers, which might be surprising given that the differences in the local star discrepancy values at the discontinuities become smaller with increasing set size (inverse linear relationship). The sampler type has relatively little importance, if any, on the performance of the optimizers. 
 We observe that uniform i.i.d. random search seems to be the best performing optimizer over the vast majority of the instances, yet still fails to come close to the exact value, as highlighted by a relative performance comparison with the known exact values. 

 We complement our empirical comparison of ``off-the-shelf'' numerical black-box optimization algorithms with a refinement of the problem-specific exact algorithm. More precisely, to stretch the settings for which we can obtain the exact $L_{\infty}$ star discrepancy values of the problem instances, we provide a new parallel implementation of the Dobkin, Eppstein, Mitchell algorithm~\cite{DEM}, the best known to this day to compute the star discrepancy. We show that our parallel implementation has a speed-up factor of up to 17 on 32 threads. This implementation is of independent interest to the discrepancy and numerical integration communities.

 \textbf{Outline.} In Section \ref{sec:definition}, we introduce the discrepancy computation problem and give some basic properties. The best discrepancy algorithms are described in Section \ref{sec:baselines}, along with a performance analysis of our parallel implementation. Section~\ref{sec:BBOsolvers} describes the instances and optimizers chosen, while Section~\ref{sec:results} describes the results obtained.

 \textbf{Availability of code.} Our implementations can be found in a Zenodo repository~\cite{reproducibiliyt}, along with past implementations by the authors of~\cite{GnewuchWW12}. The repository also contains all code and data used for the analysis of the black-box optimization algorithms, as well as the code used to process this and create the figures included in this paper.

\section{\texorpdfstring{$L_{\infty}$}{L-infinity} Star Discrepancy}
\label{sec:definition}

We provide in this section a formal definition of the $L_{\infty}$ star discrepancy and discuss some basic properties. 
\subsection{Definition and properties}
\label{sec:Props}

\begin{definition}
Let $P \subseteq [0,1)^d$ be a finite set of points. 
For $q \in [0,1]^d$, the local $L_{\infty}$ star discrepancy of $P$ in q is defined as the absolute difference between the (standard Lebesgue) volume of the box $[0,q)$, $\lambda(q)$, and the fraction of points of $P$ that fall inside this box,  
\begin{equation*} 
	d_{\infty}^*(P,q) := \left \lvert \frac{|P \cap [0,q)|}{|P|}   
		- \lambda(q) \right \rvert.
\label{eq:local}
\end{equation*}
The $L_{\infty}$ star discrepancy of $P$ measures the worst local discrepancy, i.e.,  
\begin{equation*} 
	d_{\infty}^*(P) := \sup_{q \in [0,1]^d} d_{\infty}^*(P,q).
% \label{eq:star}
\end{equation*}
\end{definition}

As mentioned in the introduction, computing the $L_{\infty}$ star discrepancy of a point set is a difficult task that likely cannot be done in polynomial time in $d$ and $|P|$. Evaluating the local $L_{\infty}$ star discrepancy, however, is rather simple and can be done naively in $O(d|P|)$. Indeed, it only requires checking if $|P|$ given points are in a specific box in $[0,1)^d$.

\begin{figure*}
    \centering
    \includegraphics[width=0.98\textwidth]{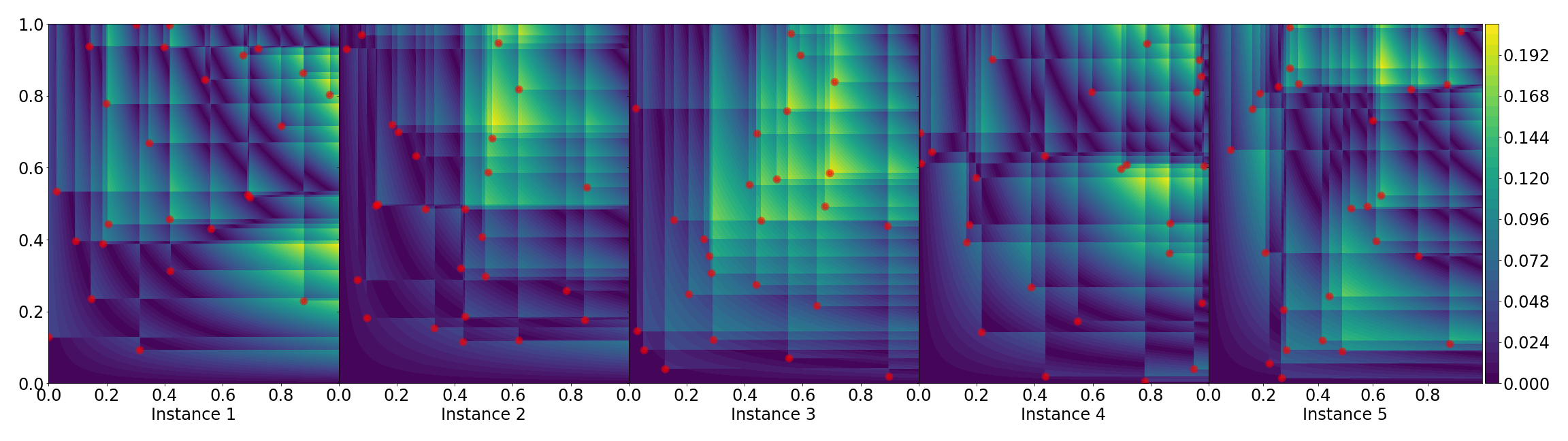}
    \caption{Discrepancy values of the first 5 instances of the 2-dimensional version of F31: Uniform sampler with 20 points. The red points indicate the originally sampled points.} 
    \label{fig:multi}
\end{figure*}

\begin{figure*}
    \centering
    \includegraphics[width=0.98\textwidth]{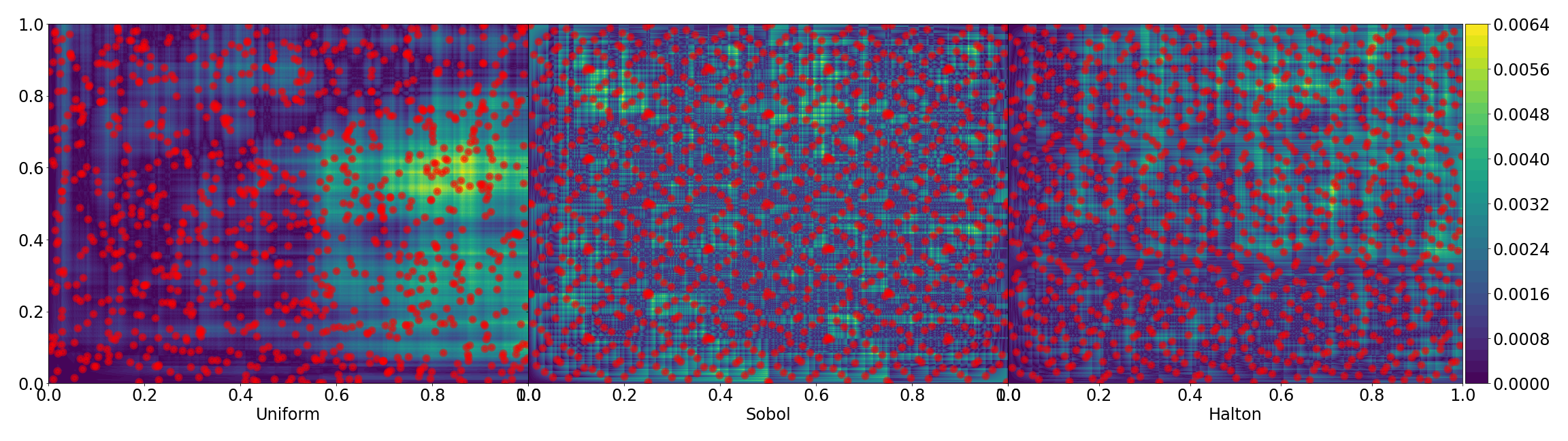}
    \caption{Comparison of the 3 samplers for 1000 points in 2D, and the corresponding discrepancy landscapes (instance 1 for F39, F49 and F59 respectively).} 
    \label{fig:grid_3samplers_1000}
\end{figure*}

\textbf{Properties of the local $L_{\infty}$ star discrepancy problem.} Figure~\ref{fig:multi} illustrates the local $L_{\infty}$ star discrepancy values for an instance in $d=2$. We can already observe that the problem of maximizing the local $L_{\infty}$ star discrepancy bears two important challenges: 
(1) it is a \textit{multimodal} problem, i.e., there can be several local optima in which the solvers can get trapped (this problem becomes worse with increasing dimension); 
(2) there are \textit{sharp discontinuities} in the local discrepancy values. Slightly increasing one coordinate can result in a point falling inside the considered box, causing a $1/|P|$ difference in the local star discrepancy value. Figure \ref{fig:grid_3samplers_1000} shows that the problem structure also depends strongly on the point set considered.

\subsection{Discrete Embedding}
\label{sec:discrete}

The star discrepancy can also be expressed as a discrete problem, as shown by Niederreiter in \cite{NieBox}. Since every closed box in $[0,1]^d$ can be obtained as the limit of a sequence of open boxes, the definition of star discrepancy can be extended to include closed boxes. We see that for a fixed $q \in [0,1]^d$, the local discrepancy can be expressed as the maximum of a local discrepancy term for the open box $[0,q)$ and a local discrepancy term for the closed box $[0,q]$, rather than with an absolute value.

More formally, we define $D(q,P)$ to be the number of points of $P:=(x^{(i)})_{i \in \{1,\ldots,n\}}$ that fall inside the open anchored box $[0,q)$, $\overline{D}(q,P)$ the number of points of P that fall inside the closed anchored box $[0,q]$ and the two following functions
\begin{equation*}
	\delta(q, P) := \lambda(q) - \frac{1}{n} D(q, P)
\qquad \text{ and } 
\qquad 
	\overline{\delta}(q, P) := \frac{1}{n} \overline{D}(q, P) - \lambda(q),
	\label{eq:delta}
\end{equation*}
where here and in the following $n=|P|$.

The local discrepancy in a point $q \in [0,1]^d$ is given by the maximum of $\delta(q,P)$ and $\overline{\delta}(q,P)$. Given this formulation, we can now show that only positions on specific grids can reach the maximum value for the star discrepancy. For all $j \in \{1,\ldots,d\}$, we define
\begin{equation*}
	\Gamma_j(P) := \{x_j^{(i)} : i \in 1,\ldots,n\} 
\qquad \text{ and } 
\qquad 
	\overline{\Gamma}_j(P) := \Gamma_j(P) \cup \{1\},
\end{equation*}
and
\begin{equation*}
	\Gamma(P) := \Gamma_1(P) \times \ldots \times \Gamma_d(P)
\qquad \text{ and } 
\qquad 
	\overline{\Gamma}(P) := \overline{\Gamma}_1(P) \times \ldots \times \overline{\Gamma}_d(P).
\end{equation*}
For any $q \in [0,1]^d \setminus \overline{\Gamma}(P)$, it is always possible to either increase or decrease slightly one of the coordinates of $q$ without changing the number of points inside the (open or closed) box. If the local discrepancy was given by $\delta(q,P)$ (respectively $\overline{\delta}(q,P)$), increasing (respectively decreasing) the volume will make the local discrepancy worse. Since only points in $\overline{\Gamma}(P)$ can reach the maximum value for the star discrepancy, we obtain with the open/closed box distinction

\begin{equation}
d_{\infty}^{*}(P) := \max \left\{\max_{q \in \overline{\Gamma}(P)}\delta(q, P),
\max_{q \in \Gamma(P)}\overline{\delta}(q, P)\right\}.
\label{discrepancy_formula}
\end{equation}

From this grid definition, the $L_{\infty}$ star discrepancy can be computed by calculating local discrepancies for each of the $O(n^d)$ points in the grid. This can be further refined by noticing that only boxes whose outer edges have at least one point on each of them (or coordinate 1 for open boxes) can reach the worst discrepancy value, bringing the total number of local discrepancies to evaluate down to $O(n^d/(d!))$. These boxes are called \emph{critical boxes}. To our knowledge, this method can be used only for relatively small sets: tests in \cite{WinkerFang} indicate that 236 points in dimension 5 or 92 in dimension 6 seem to be the limit.

 \textbf{Convention.} Since in this work we only consider the $L_{\infty}$ star discrepancy, the $L_{\infty}$ star discrepancy may be referred to as simply discrepancy or star discrepancy from hereon.

%%%%%%%%%%%%%%%%%%%%%%
\section{Problem-Specific Algorithms}
\label{sec:baselines}

We describe in this section the baselines against which we will compare the numerical black-box optimization approaches. The currently best-known \textit{exact} algorithm for computing the $L_{\infty}$ star discrepancy of a given point set is presented in Section~\ref{sec:DEM}. In Section~\ref{sec:parallelDEM} we present our parallelization of this approach. Finally, in Section~\ref{sec:TA} we briefly discuss a \textit{problem-specific local search algorithm} that we use to evaluate instances for which we cannot compute the exact value of the $L_{\infty}$ star discrepancy. 

\subsection{The DEM Algorithm}
\label{sec:DEM}

The best exact algorithm to compute the star discrepancy was introduced by Dobkin, Eppstein and Mitchell in~\cite{DEM}. It involves building a decomposition of $[0,1]^d$ in $O(n^{d/2})$ disjoint boxes $B_i$, such that we can find the maximum local discrepancy for a top-right corner in $B_i$ in linear time. This leads to a total complexity of $O(n^{1+d/2})$. As mentioned in the introduction, the algorithm can be used up to dimension 8 for a few hundred points or dimension 10 for a few dozens. While the initial algorithm in~\cite{DEM} was based on a point-box dualization, our description in this subsection will follow a more direct approach, as in Chapter 10 of \cite{DoerrGW14} and in the original implementation by the authors of \cite{GnewuchWW12}. Some border cases will be ignored for the clarity of the proof, as they do not change the general idea of the implementation or the stated complexity.

Firstly, a point $x$ is said to be \emph{internal} in dimension $j$ for a box $[a,b]$ if $a_j < x_j < b_j$. Starting from $[0,1]^d$, we build dimension-by-dimension boxes $[a,b]$ such that the two following properties are verified:
(1) Any $x \in [0,b)$ is internal in at most one dimension;
(2) For each box built up to dimension $j$ in $\{1,\ldots,d\}$, there are at most $O(\sqrt{n})$ points internal in dimension $j$ - for a fully-built box $[a,b]$ this holds for all dimensions.

The decomposition is built recursively in the following manner. We consider $[0,1]^d$ and start the decomposition in the first dimension. We find the smallest coordinate $c_{1,1}$ such that $[0,c_{1,1}] \times [0,1]^{d-1}$ contains $\sqrt{n}$ points, then $c_{1,2}$ such that $[c_{1,1},c_{1,2}] \times [0,1]^{d-1}$ contains $\sqrt{n}$ points and continue until we obtain a set $(c_{1,i})_{i \in \{0,\ldots,\lceil\sqrt{n}\rceil\}}$ where $c_{1,0}=0$ and the last non-zero $c_{1,j}=1$.% For the first dimension, $j=\lceil\sqrt{n}\rceil$, not necessarily for the next dimensions.
For each of these boxes $[c_{1,i},c_{1,i+1}] \times [0,1]^{d-1}$, we track which points are internal in dimension 1 (at most $\sqrt{n}$) and which points are inside the box $[0,c_{1,i+1}) \times [0,1]^{d-1}$.

Given a box $B_j=[c_{1,i_1},c_{1,i_1+1}] \times \ldots \times[c_{j,i_j},c_{j,i_j+1}] \times [0,1]^{d-j} $, we recursively perform a similar decomposition in dimension $j+1$. Let $n_{j}$ be the number of points inside $\overline{B_j}:=[0,c_{1,i_1+1}) \times \ldots \times[0,c_{j,i_j+1}] \times [0,1]^{d-j} $. The new $c_{j+1,i}$ need to verify the two following properties. Firstly, for any point $x$ in $\overline{B_j}$ that is internal in one of the $j$ first dimensions, there needs to be some $c_{j+1,i}=x_{j+1}$. Secondly, for any $i \in \{0,\ldots,\lceil\sqrt{n}\rceil-1\}$, there are at most $\sqrt{n}$ points $y$ in $\overline{B_j}$ such that $c_{j+1,i} < y_{j+1} < c_{j+1,i+1}$. The first property guarantees that a point will never be internal in multiple dimensions, and the second that not too many points are in the boxes obtained from $\overline{B_j}$.

After $d$ steps, we obtain boxes $[a,b]$ verifying the two desired properties: a point is internal in at most one dimension and there are at most $O(\sqrt{n})$ points internal in each dimension. We can now find the worst local discrepancy for a box whose top-right corner is in $[a,b)$ with the following dynamic programming approach. Let $m(h,j)$ (respectively $r(h,j)$) be the maximum (respectively minimum) value of $\prod_{i=1}^{j} y_i$ such that the box $[0,y_1) \times \ldots \times [0,y_j) \times [0,b_{j+1}) \times \ldots \times [0,b_d)$ contains exactly $h$ points. By ordering the points in $[0,b]$ according to their first dimension, we can easily obtain $m(\cdot,1)$ and $r(\cdot,1)$. Since points can only be internal in a single dimension, taking an internal point in dimension 1 guarantees it will be contained in the box regardless of the other choices. From the $m(\cdot,j)$ and $r(\cdot,j)$, we can therefore calculate $m(\cdot,j+1)$ and $r(\cdot,j+1)$. Calculating each $m(\cdot,j+1)$ takes $O(\sqrt{n})$ time as there are $O(\sqrt{n})$ internal points and therefore potential different choices for the coordinate $y_{j+1}$. In total, there are at most $d\sqrt{n}$ internal points, $m(\cdot,j+1)$ needs to be computed for $O(d\sqrt{n})$ different values. The dynamic programming takes $O(dn)$ time (the $d$ factor is usually ignored). All that remains to be done is to find $\max_{1 \leq h\leq n}(m(h,d)-h/n,h/n-r(h,d))$, in other words which (number of points inside a box, box volume) combination gives the worst discrepancy. This will directly give the worst discrepancy value for a box whose top-right corner is in $[a,b)$.

%%%%
\subsection{Parallelizing DEM}
\label{sec:parallelDEM}

To parallelize the DEM algorithm we start by noting that after
computing the boxes for a given dimension we can continue the
recursive decomposition independently for each box. This naturally
gives way to a parallel-task construct where each task that can be
scheduled to an available CPU corresponds to the decomposition in dimension $j+1$ of a
given box $B_j$.

At this point, it is worth noting that another option
initially considered was to parallelize the dynamic programming
algorithm that is used to compute the worst local discrepancy for a
box after $d$ steps. However, this idea was discarded since the
dynamic programming algorithm was very fast in practice and
preliminary attempts did not yield significant speedups.

To implement the parallel-tasks construct we considered OpenMP
Tasks~\cite{ayguade2009design} and we adapted the implementation
of the DEM algorithm by the authors of \cite{GnewuchWW12}. In particular, one relevant change was
that the sequential implementation reused the set of points between recursive
calls, with a possible resorting of the points in the current
dimension after each recursive call. However, in the parallel variant,
we cannot efficiently share the set of points since sorting operations
on the set of points in a thread could potentially affect tasks running
in other threads. Instead, in the parallel DEM we opted to copy the set
of points before entering a recursive call.
Although this requires an extra copy before each recursive call, we observed
that in some cases copying the vector was faster than resorting after each
recursive call, suggesting that the non-parallel DEM implementation could
be slightly improved by taking this into account.

Finally, in order to avoid thread starvation with a recursive parallel-tasks
construct, it is often useful to consider a cut-off value related to
the problem size to switch to a non-parallel variant. In this case,
we define the cut-off with respect to the complexity
of the DEM algorithm, i.e., $O(n^{1+d/2})$. In particular, when
decomposing a box $B_j$ with $n_j$ points in dimension $j$, we switch to
the non-parallel DEM algorithm if ${n_j}^{1+(d-j)/2} < 1e8$. This
cut-off value was defined empirically by experimenting with different
values up to $1e15$.

Table~\ref{tab:parallel_times} gives the CPU time in seconds for the
parallelized DEM on several instances generated with the GNU Scientific
Library (using the indicated sequence) and considering different numbers of
threads. For a single thread, the best time between the non-parallel and parallel
implementations is shown. Experiments were run on a machine with two
Intel(R) Xeon(R) Silver 4210R CPUs clocked at 2.40GHz, comprising a
total of 20 cores and 40 threads. We can see that for 32 threads we
have speedups that range from a factor of 7.4 to a factor of 17.4.
These results show that parallelization is generally successful and
can give a significant boost. However, we believe that there is still room
for improvement by further tuning the cut-off between the parallel and
non-parallel variants, and by further analyzing when it is best to copy
or resort the set of points. 

\begin{table*}[]
    \centering
    % latex table generated in R 4.2.1 by xtable 1.8-4 package
\begin{tabular}{lllllllll}
  \toprule
  \multicolumn{3}{c}{} & \multicolumn{6}{c}{\textbf{CPU time (speedup) per number of threads}}\\
\cmidrule(l){4-9}
N. Dim. & N. Points & Sequence & 1 & 2 & 4 & 8 & 16 & 32\\
 \midrule
\midrule\multirow{4}{*}{\textbf{2}} & \multirow{4}{*}{\textbf{50000}} & halton & 2.54 (1.0) & 1.56 (1.6) & 0.88 (2.9) & 0.53 (4.8) & 0.38 (6.7) & 0.32 (7.9) \\ 
   &  & niederreiter & 2.50 (1.0) & 1.56 (1.6) & 0.87 (2.9) & 0.54 (4.6) & 0.34 (7.4) & 0.31 (8.1) \\ 
   &  & reversehalton & 2.37 (1.0) & 1.59 (1.5) & 0.86 (2.8) & 0.53 (4.5) & 0.40 (5.9) & 0.32 (7.4) \\ 
   &  & sobol & 2.39 (1.0) & 1.58 (1.5) & 0.86 (2.8) & 0.53 (4.5) & 0.41 (5.8) & 0.31 (7.7) \\ 
  \midrule\multirow{4}{*}{\textbf{3}} & \multirow{4}{*}{\textbf{10000}} & halton & 15.70 (1.0) & 8.19 (1.9) & 4.31 (3.6) & 2.37 (6.6) & 1.52 (10.3) & 1.13 (13.9) \\ 
   &  & niederreiter & 15.77 (1.0) & 8.09 (1.9) & 4.36 (3.6) & 2.39 (6.6) & 1.65 (9.6) & 1.16 (13.6) \\ 
   &  & reversehalton & 15.57 (1.0) & 7.99 (1.9) & 4.37 (3.6) & 2.37 (6.6) & 1.54 (10.1) & 1.17 (13.3) \\ 
   &  & sobol & 15.74 (1.0) & 8.19 (1.9) & 4.38 (3.6) & 2.39 (6.6) & 1.44 (10.9) & 1.14 (13.8) \\ 
  \midrule\multirow{4}{*}{\textbf{4}} & \multirow{4}{*}{\textbf{3000}} & halton & 49.66 (1.0) & 26.40 (1.9) & 13.97 (3.6) & 7.38 (6.7) & 3.99 (12.4) & 2.89 (17.2) \\ 
   &  & niederreiter & 50.31 (1.0) & 26.82 (1.9) & 14.33 (3.5) & 7.42 (6.8) & 3.96 (12.7) & 2.89 (17.4) \\ 
   &  & reversehalton & 50.42 (1.0) & 26.53 (1.9) & 14.11 (3.6) & 7.44 (6.8) & 4.05 (12.4) & 2.96 (17.0) \\ 
   &  & sobol & 50.41 (1.0) & 27.20 (1.9) & 14.31 (3.5) & 7.32 (6.9) & 4.06 (12.4) & 2.92 (17.3) \\ 
  \midrule\multirow{4}{*}{\textbf{5}} & \multirow{4}{*}{\textbf{1000}} & halton & 58.87 (1.0) & 32.54 (1.8) & 16.70 (3.5) & 9.18 (6.4) & 5.10 (11.5) & 3.83 (15.4) \\ 
   &  & niederreiter & 62.32 (1.0) & 33.39 (1.9) & 17.78 (3.5) & 9.67 (6.4) & 5.34 (11.7) & 4.08 (15.3) \\ 
   &  & reversehalton & 62.08 (1.0) & 32.77 (1.9) & 17.54 (3.5) & 9.64 (6.4) & 6.88 (9.0) & 3.90 (15.9) \\ 
   &  & sobol & 63.80 (1.0) & 33.89 (1.9) & 17.73 (3.6) & 9.77 (6.5) & 5.34 (11.9) & 4.04 (15.8) \\ 
  \midrule\multirow{4}{*}{\textbf{6}} & \multirow{4}{*}{\textbf{600}} & halton & 175.79 (1.0) & 103.74 (1.7) & 53.04 (3.3) & 28.82 (6.1) & 17.12 (10.3) & 12.57 (14.0) \\ 
   &  & niederreiter & 196.24 (1.0) & 109.22 (1.8) & 57.99 (3.4) & 30.44 (6.4) & 17.94 (10.9) & 12.71 (15.4) \\ 
   &  & reversehalton & 187.99 (1.0) & 114.35 (1.6) & 58.75 (3.2) & 31.56 (6.0) & 19.12 (9.8) & 13.49 (13.9) \\ 
   &  & sobol & 200.76 (1.0) & 112.53 (1.8) & 59.18 (3.4) & 30.91 (6.5) & 18.34 (10.9) & 13.32 (15.1) \\ 
  \midrule\multirow{4}{*}{\textbf{7}} & \multirow{4}{*}{\textbf{300}} & halton & 160.25 (1.0) & 94.43 (1.7) & 48.22 (3.3) & 27.10 (5.9) & 15.64 (10.2) & 11.70 (13.7) \\ 
   &  & niederreiter & 167.32 (1.0) & 96.85 (1.7) & 51.36 (3.3) & 28.40 (5.9) & 15.80 (10.6) & 12.66 (13.2) \\ 
   &  & reversehalton & 149.43 (1.0) & 88.79 (1.7) & 50.57 (3.0) & 29.72 (5.0) & 16.63 (9.0) & 12.49 (12.0) \\ 
   &  & sobol & 169.70 (1.0) & 102.12 (1.7) & 52.72 (3.2) & 29.68 (5.7) & 16.20 (10.5) & 12.41 (13.7) \\ 
   \bottomrule
\end{tabular}

    \caption{CPU time in seconds and corresponding speedup in parenthesis for the parallelized DEM algorithm}
    \label{tab:parallel_times}
\end{table*}

%%%%
\subsection{The TA Algorithm}
\label{sec:TA}

Several applications of point sets with low star discrepancy value concern settings that are not efficiently tractable by the exact algorithms described in the previous two subsections. For these applications, we therefore need to resort to heuristic approaches to evaluate the discrepancy of a given point set. To date, the best-known heuristic is a Threshold Accepting (TA) algorithm proposed in~\cite{GnewuchWW12}. This approach exploits the grid structure introduced in Section~\ref{sec:discrete} and operates on the search space $[1..n+1]^d$, with each point encoding one of the grid points in $\overline{\Gamma}(P)$. 

The TA algorithm from~\cite{GnewuchWW12} builds on an earlier approach by Winker and Fang suggested in~\cite{WinkerFang} and extends it by various problem-specific components. Threshold Accepting~\cite{TADueck} is similar to Simulated Annealing~\cite{SA83} but replaces its probabilistic selection criterion with a deterministic one. That is, at each step, the current incumbent solution is compared to a randomly sampled neighboring solution. The neighbor is selected as the new center if its quality is not much worse than that of the previous incumbent. More precisely, neighbor $y$ replaces incumbent $x$ if $f(y)-f(x)\ge \tau(t)$, where  $\tau(t) \le 0$ is the threshold chosen at iteration $t$. The sequence $(\tau(t))_{t}$ is monotonically increasing so that the further advanced the optimization process is, the harder the selective pressure. 

The TA algorithm from~\cite{GnewuchWW12} uses a dynamic choice of the neighborhood structure, increasing the number of coordinates $\{j \mid x_j \neq y_j\}$ that may change in each iteration while at the same time decreasing the absolute difference $|x_j-y_j|$. The more important problem-specific component, however, is a ``snapping'' routine, which rounds a selected grid point to a so-called critical box; see~\cite{GnewuchWW12} for details. 

Using the DEM solver from Section~\ref{sec:DEM} as a baseline, it was shown in~\cite{GnewuchWW12} that the TA algorithm successfully found the optimum on all instances for which DEM could provide exact values. Based on the results presented in~\cite{GnewuchWW12}, the TA algorithm seems reliable for point sets up to dimensions 12 to 20 for a few hundred points. 
% A more detailed description of these algorithms and other attempts to compute the star discrepancy can be found in Chapter 10 in \cite{DGWBook}.

%%%%%%%%%%%%%%%%%%%%%%
\section{Numerical Black-Box Optimization Approaches}
\label{sec:BBOsolvers}

% \subsection{Experimental Setup}

The $L_\infty$ star discrepancy problem was included as a black-box benchmark problem in IOHexperimenter (version 0.3.7)~\cite{IOHexperimenter}, using three different point set generators: 
\begin{enumerate}
    \item uniform random sampling, 
    \item Halton~\cite{Halton60},  
    \item Sobol'~\cite{Sobol}.
\end{enumerate}
 For the uniform sampler, a standard Mersenne Twister 19937 pseudo-random number generator was used, provided by the C++ STL. The Halton sequence was generated using a classic Sieve of Eratosthenes prime number generating algorithm, and the Sobol' sequence was generated with a third-party library based on the FORTRAN implementation of~\citet{i8sobol}. 
 
In addition to the generator, the number of points to be sampled can be selected, which along with the dimensionality, controls the complexity of the problem. We define a default suite, which includes for every generator a fixed number of samples $S\in\{10,25,50,100,150,200,250,500,750,1000\}$. This includes a total of $30$ benchmark problems, where for each problem the dimension and instance can be varied arbitrarily. Instances are controlled by a unique instance identifier, which is a positive integer that determines the random seed used to generate the point set. The problems can be accessed through both the Python and C++ interfaces of IOHexperimenter, with problem ids $\{30,\dots, 39\}$, $\{40,\dots, 49\}$, and $\{50,\dots, 59\}$ for the problems generated with the uniform, Sobol', and Halton generators, respectively.

We test a total number of eight algorithms, all taken from the Nevergrad platform~\cite{nevergrad}: 
\begin{enumerate}
    \item Diagonal Covariance Matrix Adaptation Evolution Strategy (dCMA-ES)~\cite{hansen2001self_adaptation_es}
    \item NGOpt14, Nevergrad's algorithm selection wizard~\cite{meunier2021black}
    \item Estimation of Multivariate Normal Algorithm (EMNA)~\cite{emna}
    \item Differential Evolution~\cite{StornPriceDE}
    \item Constrained Optimization BY Linear Approximation (Cobyla)~\cite{cobyla}
    \item Random Search
    \item Particle Swarm Optimization (PSO)~\cite{PSO}
    \item Simultaneous Perturbation Stochastic Approximation algorithm (SPSA)~\cite{spsa}
\end{enumerate}
The algorithms are chosen ``as they are" from Nevergrad. That is, we did not perform any hyper-parameter tuning nor did we change any of their components. 
Each algorithm is given a total budget of $2\,500 \cdot d$ local discrepancy evaluations, where $d$ is the dimensionality of the problem. For each instance, 10 independent runs of the algorithm are performed. This is repeated for 10 instances, resulting in a total of 100 runs per function, for each of the $30$ functions, with dimensionality $d\in\{2,3,4,6,8,10,15\}$.

We run all our experiments in the IOHprofiler environment~\cite{IOHprofiler}. This allows us to track not only the final performance, but also the trajectory of the algorithms in objective space. It furthermore allows a straightforward visualization and analysis of the data using the IOHanalyzer module~\cite{IOHanalyzer}. 

For all instances in dimensions 2, 3 and 4, and for all the instances in dimensions 6, 8, 10, and 15 with $n$ not larger than 750, 200, 50 and 10 points respectively, we computed the exact discrepancy values using the parallel DEM algorithm proposed in Section~\ref{sec:parallelDEM}. For all other instances, we computed a lower bound for the star discrepancy value using the TA algorithm described in Section~\ref{sec:TA}. 

%%%%%%%%%%%%%%%%%%%%%%
\section{Results}
\label{sec:results}

\begin{figure}
    \centering
    \includegraphics[width=0.48\textwidth]{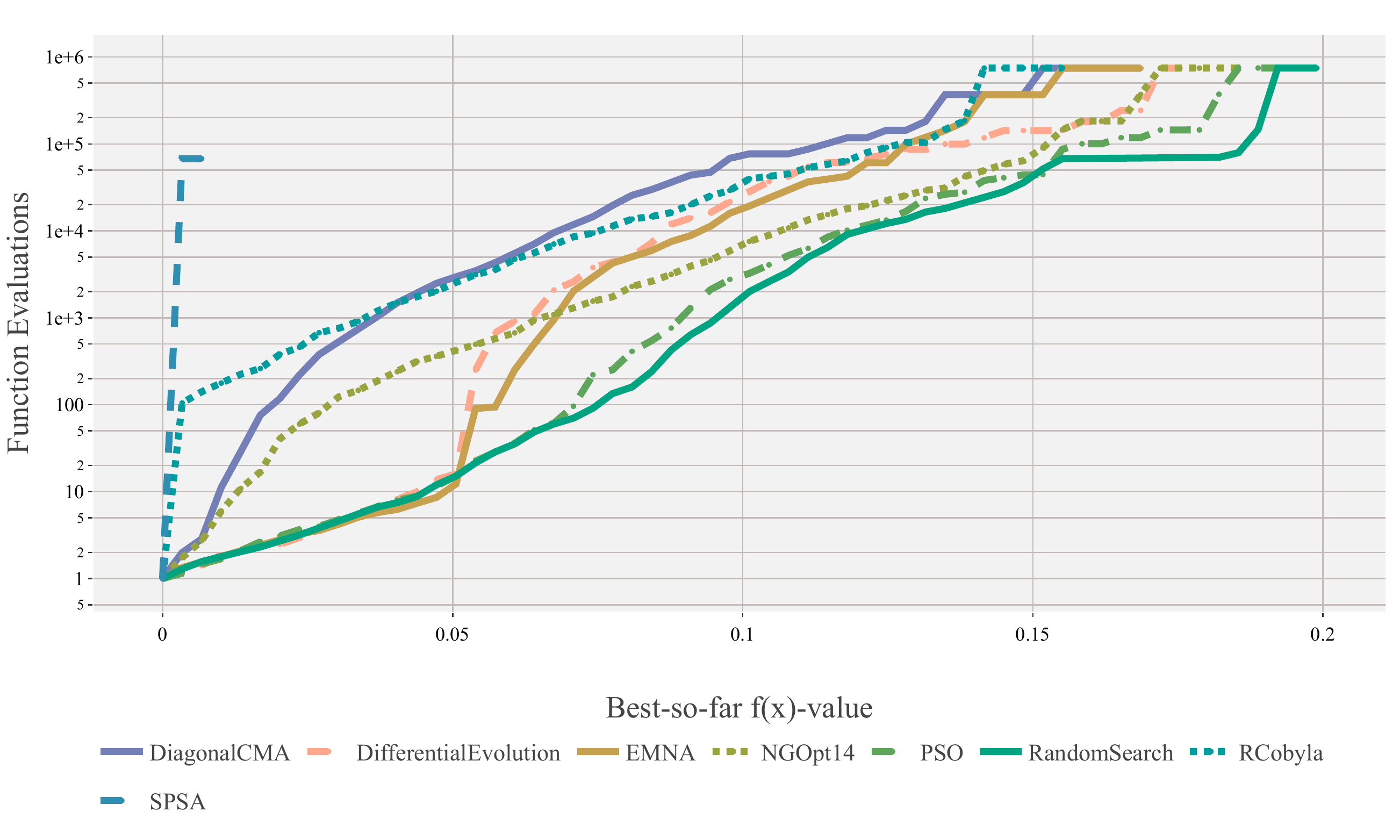}
    \caption{Expected running time (ERT) of the 8 optimization algorithms on the discrepancy calculation for the uniform sampler with 100 samples (F33) in 3 dimensions. ERT is calculated based on 10 runs on 10 instances with a budget of $7\,500$. Figure generated using IOHanalyzer~\cite{IOHanalyzer}.}
    \label{fig:ert_example}
\end{figure}

We can compare the performance of the used optimization algorithms by considering the expected running time (ERT) to reach increasing discrepancy values. This analysis shows the convergence behavior on the selected function. In Figure~\ref{fig:ert_example}, we show the ERT on the 3-dimensional version of F33: the uniform sampler with $n=100$.  From this figure, we can clearly see that the algorithms struggle to optimize this function. Particularly noticeable is the SPSA algorithm, which seems to fail to find even slightly improved discrepancy values. On the other side, we notice that Random Search is surprisingly outperforming all other optimizers. This seems to indicate that even for this relatively simple setting of $n=100$ and $d=3$, the high level of multi-modality combined with discontinuities in the landscape cause problems for all of the considered optimization algorithms.

To check whether this is a consistent problem, or something specific to the settings chosen in Figure~\ref{fig:ert_example}, we can look in more detail at the final solutions found by each optimizer across a wider set of scenarios. In order to create a fair comparison, we can move from the original discrepancy values to a relative measure, based on the bounds found by the TA and DEM algorithms. Specifically, we consider the following measure:
$$R(x)= \frac{OPT(x) - f(x)}{OPT(x)},$$
where $f(x)$ is the final value found after the optimization run, and $OPT(x)$ is the bound calculated by the parallel DEM or TA, depending on the instance size. 

\begin{figure*}
    \centering
    \includegraphics[width=0.98\textwidth]{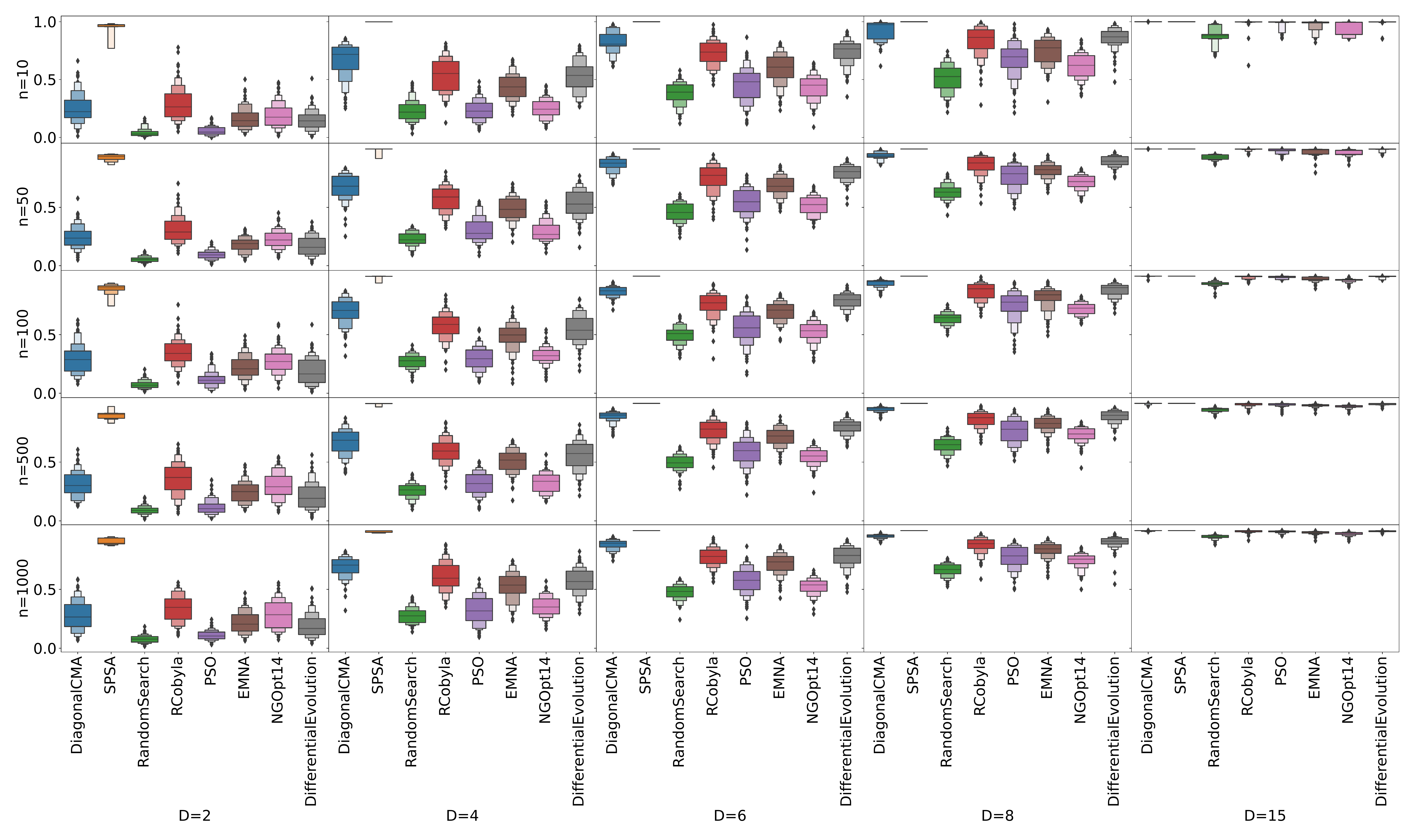}
    \caption{Relative final discrepancy value found by each of the used optimizers. Values of 0 correspond to finding the optimal solution, while 1 corresponds to the worst achievable value (0 discrepancy). Box-plots are aggregations of 10 runs on 10 instances, all for the uniform sampler. }
    \label{fig:grid_boxplots_unif_smaller}
\end{figure*}

Using this relative measure, we can compare the final solutions found by each optimization algorithm across a set of different $n$ and $d$. This is visualized in Figure~\ref{fig:grid_boxplots_unif_smaller}. In this figure, we see that the observations made based on ERT from Figure~\ref{fig:ert_example} seem to hold across scenarios: the SPSA algorithm is clearly performing poorly, while Random Search seems to be competitive with, if not superior to, all other algorithms for every scenario. In addition to the ranking between algorithms, we also note a clear increase in problem difficulty as the dimensionality increases. Conversely, the number of samples seems to have a rather limited impact on the relative difficulty. This suggests that the structure of the point set has little influence on the performance of the optimizers.

\begin{figure*}
    \centering
    \includegraphics[width=0.98\textwidth]{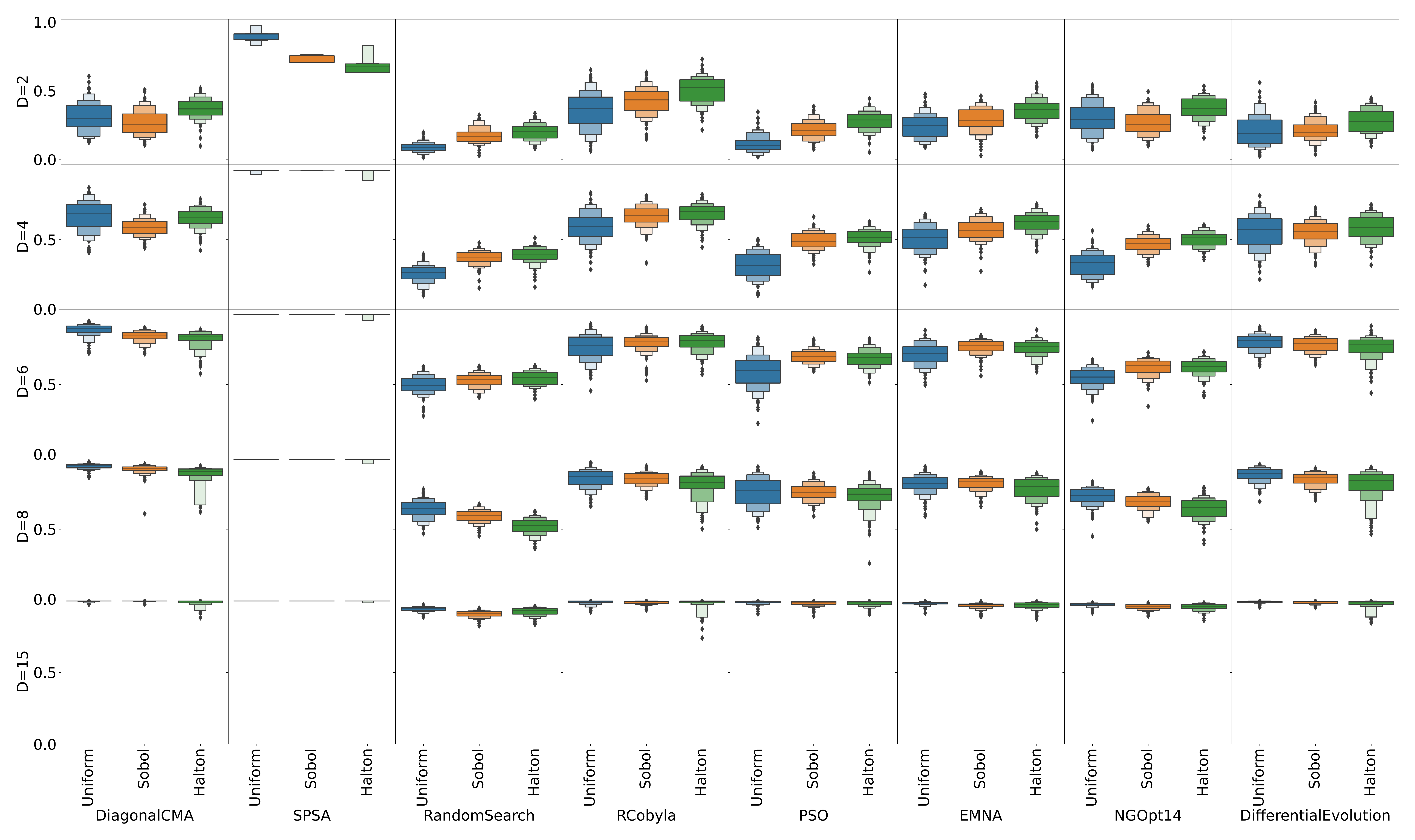}
    \caption{Relative final discrepancy value found by each of the used optimizers, compared between different samplers. Values of 0 correspond to finding the optimal solution, while 1 corresponds to the worst achievable value (0 discrepancy). Boxplots are aggregations of 10 runs on 10 instances, all for $n=500$. }
    \label{fig:grid_across_samplers}
\end{figure*}

As a final comparison, we can consider the differences in the difficulty of the optimization problem when different samplers are used. As we observed in Figure~\ref{fig:grid_3samplers_1000}, the landscape is clearly impacted by the choice of the sampler. To see whether this also impacts the performance of the optimization algorithms, we consider the relative final discrepancy found on the grids with $n=500$, for a few selected dimensions. The resulting distributions are visualized in Figure~\ref{fig:grid_across_samplers}. From this figure, we can see that, while the choice of sampler often has a low impact on the performance of most algorithms, some tendencies can still be observed. In most cases, the Halton sampler seems to be the most challenging, especially in lower dimensions. However, the ordering is not fully consistent between algorithms or even between dimensions. 

%%%%%%%%%%%%%%%%%%%%%%
\section{Conclusions}
We have studied the efficiency of numerical black-box optimization approaches for maximizing the local $L_{\infty}$ star discrepancy values for a given point set $P$. The results are underwhelming; the obtained results cannot be used as reliable estimates for the overall $L_{\infty}$ star discrepancy of $P$. 

The results indicate that off-the-shelf black-box optimization approaches have difficulties coping with the multi-modal nature of the problem and/or with the discontinuities in the genotype-phenotype mapping. We believe that this combination makes the problem an interesting use case for comparing diversity mechanisms with or without restarts. In particular, we expect that approaches such as \textit{quality-diversity} (originally introduced in~\cite{PughSSS15,PughSS16}, but see \cite{QDChatzi} for a more recent survey) or \emph{niching}~\cite{Shir2012niching} could improve the quality of the search algorithms. 

The focus of our work has been on the numerical black-box solvers. However, an interesting aspect of the star discrepancy computation problem is that it can also be studied as a discrete problem, using the grid structure described in Section~\ref{sec:definition}. This grid structure is exploited by the TA algorithm from~\cite{GnewuchWW12} (see Section~\ref{sec:TA}). We suspect that merging some of the problem-specific components of this algorithm (e.g., the \textit{snapping} rounding routines) into evolutionary algorithms operating on discrete search spaces of the type $[1..n]^d$ could be worthwhile. 

As we have motivated in the introduction, computing the $L_{\infty}$ star discrepancy of a given point set is crucial for the design of low-discrepancy point sets, which have numerous applications in a broad range of industrial and academic problems, including evolutionary computation~\cite{NeumannGDN018,BossekDKNN20}. Apart from providing a challenging configurable problem suite for benchmarking black-box optimization approaches, the design of problem-specific evolutionary strategies or similar approaches would be highly desirable. Both the discrepancy community, via a better understanding of the structure of the discrepancy function, and any application relying on numerical integration or Quasi-Monte Carlo integration would benefit greatly from effective solvers for this problem.

%%%%%%%%%%%%%%%%%%%%%%
\begin{acks}
Our work is financially supported by ANR-22-ERCS-0003-01 project VARIATION, by the CNRS INS2I project IOHprofiler, and by Campus France Pessoa project DISCREPANCY.
This work is partially funded by the FCT - Foundation for Science and
Technology, I.P./MCTES through national funds (PIDDAC), within the
scope of CISUC R\&D Unit -- UIDB/00326/2020 or project code
UIDP/00326/2020.
\end{acks}

% \bibliographystyle{alpha}
% \bibliography{references}%,bib/abbrev,bib/journals,bib/authors,bib/biblio,bib/crossref} 
\newcommand{\etalchar}[1]{$^{#1}$}

\end{document}